\documentclass[sigconf, screen, balance]{acmart}

\usepackage{graphicx}

\usepackage{xspace}
\usepackage{algorithm}
\usepackage{algorithmic}
\usepackage{enumitem}
\usepackage{enumerate}
\usepackage{CJKutf8}
\usepackage{CJK}
\usepackage{comment}
\usepackage{times}
\usepackage{xcolor}
\usepackage{appendix}
\usepackage{url}
\definecolor{lightgray}{rgb}{0.7,0.7,0.7}
\usepackage[]{lineno}

\usepackage[normalem]{ulem}

\newcommand{\kk}{\textsc{40}\xspace}
\newcommand{\hide}[1]{}

\newcommand{\lastcut}[1]{\textcolor{black}{#1}} 
\newcommand{\revise}[1]{{\color{black} #1}}

\newcommand{\auc}{AUC${}_0$\xspace}
\newcommand{\acc}{ACC${}_0$\xspace}

\newcommand{\aucww}{AUC\xspace}
\newcommand{\accww}{ACC\xspace}
\newcommand{\nblr}{NBLR\xspace}
\newcommand{\nblrw}{\revise{CSLR${}_w$\xspace}}

\newcommand{\yifan}[1]{{\color{black}#1}}

\newcommand{\ssa}{\textsc{SSA Data}\xspace}
\newcommand{\xobni}{\textsc{Yahoo Data}\xspace}
\newcommand{\xobniB}{\textsc{Yahoo Full Name Data}\xspace}

\newcommand{\andrea}{\textsc{Andrea Data}\xspace}
\newcommand{\toni}{\textsc{Toni Data}\xspace}
\acmConference[]{}{}{}
\begin{document}

\title{What's in a Name? -- Gender Classification of Names with Character Based Machine Learning Models}

\author{Yifan Hu${}^1$, Changwei Hu${}^1$, Thanh Tran${}^1$,  Tejaswi Kasturi${}^1$, Elizabeth Joseph${}^2$, Matt Gillingham${}^1$}

\affiliation{
  \institution{${}^1$Yahoo! Research, ${}^2$Verizon Media}
  \institution{${}^2$Worcester Polytechnic Institute, Worcester, MA, USA}
}

\email{yifanhu@verizonmedia.com}

\begin{abstract}
Gender information is no longer a mandatory input when registering for an account at many leading Internet companies.
However, prediction of demographic information such as gender and age remains an important task, especially in intervention of unintentional gender/age bias in recommender systems.
Therefore it is necessary to infer
the gender of those users who did not
to provide this information during registration.
We consider the problem of predicting the gender of registered
users based on their declared name.
By analyzing the first names of 100M+ users, we found that
genders can be very effectively classified using the composition of the name strings. We propose a number of character based machine learning models, and demonstrate that our models are able to infer the gender of users with
much higher accuracy than baseline models. \revise{Moreover, we show that using the last names 
in addition to the first names improves classification performance further.}
\end{abstract}

\maketitle

\section{Introduction}
\label{sec:intro}
Gender information is no longer a required input when registering for an account at many leading Internet companies. Consequently, a significant
percentage of registered users do not have a declared gender; and over time 
this percentage is expected to rise.  \lastcut{In many tasks, such as when checking for fairness in
machine learning algorithms for content recommendation \cite{karako2018using,yao2017beyond}}, 
it is important to know users' demographic information, of which
gender is an integral part.  Therefore, it is necessary to
infer the gender of those registered users who do not disclose
it, by using information collected during their
registration, and/or from their subsequent activities. 
Although we are aware that gender identity is not
necessarily binary, we treat gender inference as a binary classification
problem in this paper.  In addition, this research was conducted in accordance with Verizon Media's Privacy Policy and respect for applicable user privacy controls.

One possible way to infer gender is to examine user activities. We experimented with a content based model, where features are extracted from the content the users interacted with through activities such as browsing.
The content is distilled into categorical features including wiki-entities (i.e., entries
in Wikipedia), 
and proprietary content categories derived via ML models (similar to content categories at \cite{google_categories}). These categorical features are used in a logistic regression
algorithm to classify user genders. The advantage of such \yifan{ a content-based model} is that it identifies
users by their browser's cookies, thus \yifan{it works} even for users who are
unregistered or not logged in. However, a limitation of \yifan{such} content-based approaches is that 
signals for users can be sparse. Additionally, signals may be lost in
the process of classifying content into categorical features. As a
result, we found that the performance of this content-based approach is modest.

Therefore we are interested in enhancing the above baseline model 
for registered users, for whom we have more information.   
In particular, we believe that the user's name
 could be used to infer gender effectively. Accordingly, we propose a number of character-based
machine learning approaches for the gender classification task. Our experimental results on 100M+ users demonstrate that the
name-based approach improves AUC (of ROC) by $0.14$ compared with the content-based model.

The major contributions of our work are:
\begin{itemize}

    \item  We propose several character-based
     techniques to predict users' genders based on their first name. By constructing both
     a linear model and various deep neural models, we show that gender can be predicted very effectively using information encoded in the spelling of the user's first name. 

     \item{} We find surprisingly that a linear model, combined with judicious feature engineering, performed just as well as complex models such as character-based LSTM~\cite{lstm_hoch_1997} or BERT~\cite{devlin2018bert}, and better than other character-based deep learning models, an embedding-based model, and a baseline content-based model.

     \item{} We conduct our experiment on a very large real-world dataset of 21M unique first names and show that
     our models trained on this large dataset extend better than the same models trained on a medium sized public dataset.
     We also show that our models perform well on unseen or multilingual datasets.
     
	\item \revise{We demonstrate that the performance could be further enhanced when utilizing both the first and  the last names through our proposed dual-LSTM algorithm. We show that fusing name and content information can also boost performance.}
	
\end{itemize}

\revise{Throughout the paper, we shall denote $P(M)$/$P(F)$  the probability of male/female, and 
$P(M|X)$  the probability of male given feature set $X$. The goal of the proposed 
machine learning models is to compute $P(M|X)$, 
based on the user feature set $X$.
All models in this paper minimize the binary entropy loss function.  
}

The outline of this paper is as follows.
Section~\ref{sec:data} introduces datasets used for our models. Section~\ref{sec:models} proposes numerous character-based models for gender classification.
Section~\ref{sec:results} compares performance of these models.
\revise{Section~\ref{sec:fullname} utilizes both the first and last names to enhance performance.}
Section~\ref{sec:related} reviews related work.
We conclude in Section~\ref{sec:conc} with
discussions on limitations and future work. 
\revise{In the Appendix, we 
propose a fusion model  utilizing both name and content, and discuss  
details of using the gender prediction models in practice, as well as how to
understand the models.}

\section{Data and the Baseline Model}
\label{sec:data}
To infer genders of registered users, at least two kinds of data can be used. One is the content that the users interacted with, and the other is the declared information of the users during registration, particularly names. The former 
is the premise for a content-based baseline model we tested. However we believe that the latter could be an important signal for inferring the gender.
Our work is based on two datasets, one from the Social Security Administration (SSA), and the other  from Verizon Media. Throughout this paper, when we talk about the name of a person, we are referring to the first name, unless otherwise specified.

\subsection{SSA baby names dataset}

The Social Security Administration collects data of names, genders, and name frequencies ~\cite{ssa_babies} for newborn babies in the US, spanning from 1880 to 2018 (139 years). 
For babies born 
in 2018, the top 4 most popular names are ``Olivia (female, frequency: 17,921); Noah (male, frequency: 18,267), Emma (female, frequency: 18,688), and Liam (male,frequency: 19,837)''. Around 11\% of the names appear both as male and female. For example, in 2018, the name ``Avery'' is documented as both female (8,053 times) and male (2,098 times).
Henceforth, we will call this dataset \ssa. 
When aggregating over all 139 years, we found 98,400 unique names from a total of 
351,653,025 babies, 
of which 50.5\% are \revise{boys}. 
Among the 98,400 unique names, 10,773 of them are used by both  \revise{boys} and \revise{girls}.
Table~\ref{tab:ssa_xobni_top}~(left) shows the top 5 male and female names.

\begin{table*}[t]
    \centering
        \setlength{\tabcolsep}{4.5pt}
    \begin{tabular}{c|c|c|c|c|c|c|c}
\multicolumn{4}{c|}{\ssa} & \multicolumn{4}{c}{\xobni}\\
\hline
male name & count & female name & count & male name & count/$p(M)$ & female name & count/$p(M)$\\
\hline
James & 5.2M & Mary & 4.1M & John & 2.4M/0.977 & Maria & 1.0M/0.031\\
John & 5.1M & Elizabeth & 1.6M & David & 2.1M/0.981 & Mary & 0.9M/0.054 \\
Robert & 4.2M & Patricia & 1.6M & Michael & 1.9M/0.983 & Jennifer & 0.9M/0.027\\
Michael & 4.4M & Jennifer & 1.5M & James & 1.7M/0.974 & Jessica & 0.8M/0.021\\
William & 4.1M & Linda & 1.5M & Robert & 1.4M/0.979 & Sarah & 0.8M/0.025\\
    \end{tabular}
    \caption{Left: top male and female US baby names in \ssa. Right: top male and female names in \xobni. Here $p(M)$ is the observed probability of the name being male.}
    \label{tab:ssa_xobni_top}
\end{table*}

\subsection{Verizon Media dataset}

This dataset is a sampled dataset from Verizon Media (formerly Yahoo! and AOL), which contains a total of 100M+ of users. To be consistent with \ssa, we aggregate by first name to give 
us the final dataset,
which contains ``first name, gender, \#male, \#female.'' Here ``first name'' can contain one or more tokens, such as ``John Robert''.
We assign a label of 1 if there are more \revise{men} using this first name than \revise{women}, and 0 otherwise \revise{(please see Section~\ref{sec:results}
for a probabilistic treatment of labels)}. This dataset, called \xobni from now on, contains 21M unique first names. They follow 
a typical power law distribution with a long tail. Over
72\% of names only appear once, 10\% appear twice, and 
4\% appear 3 times. On the other end of the scale,
the most popular names are shown in Table~\ref{tab:ssa_xobni_top}~(right). Compared with \ssa, the popular names are different. This reflects both the fact that Verizon Media users 
are international (for example, ``Maria'' is a popular female name in Spanish speaking countries), and the fact that the popularity of names changes over time and that the \ssa covers 139 years. \lastcut{In addition, \xobni is  less certain compared with \ssa. For example
in Table~\ref{tab:ssa_xobni_top}, top names are around 97.0\% male or female, while these same names are recorded as 100\% male or female in \ssa.}

\begin{table}[thb]
    \centering
    \begin{tabular}{c|c|c|c}
k & name count & Female Coverage & Male Coverage\\
\hline
5&3.1M&92.65&92.43\\
10&1.7M&90.84&90.60\\
20&1.0M&88.96&88.61\\
40&0.5M&86.93&86.29\\
80&0.3M &84.74&83.85\\
100&0.3M&83.59&82.71\\
\end{tabular}
    \caption{Percentage of users in \xobni  with name frequency $\ge k$.}
    \label{tab:freq_coverage}
\end{table}

Table~\ref{tab:freq_coverage} shows the ratios of users that are covered by ``popular" names. If
we take all 0.3M names that appear at least $100$ times in the data, 84\% of \revise{men} and 83\% of \revise{women} are covered.
If we take all the 3.1M names that appears at least $5$ times, 
92\% of users are covered.

Note that the size of \revise{the set of} ``popular" names in \xobni
is considerably larger than the whole \ssa. Even imposing a high frequency of 100 or more, we end up with 
300K names, 
\emph{vs.} 
98K names in the \ssa. 
One reason for the larger size of \revise{the set of} popular names is
that the names in \xobni may contain multiple tokens. About 34\% of names contain a mixture of \revise{letters} and spaces, e.g. ``Ana Carolina'' or ``Carlos Eduardo''. Another reason is that 
Verizon Media has a substantial international presence. 
A check on a random sample of \xobni shows that about 2.2\% of the
names contain non-English characters. The most popular of such names include ``Andr\'e'' and ``Björn'' for \revise{men}, and ``Aur\'elie'' and ``B\'arbara'' for \revise{women}. Moreover, around 11\% of first names contain \revise{non-letters}, such as 
``Anne\_Marie", ``Asma'a".

Since we are interested in inferring gender for Verizon Media users, and given the significant differences between the \ssa and \xobni, we propose to train a number of name-based machine learning models using \xobni, but shall verify our models using both \ssa and \xobni.

\revise{
Finally, in order to study the effect of using full names  (Section~\ref{sec:fullname}), we randomly sampled 
13M users, each with a full name, 
together with the declared genders, out of the 100M+ users in the \xobni. We call this set \xobniB.
}

\subsection{Content data for the baseline model\label{sec:content}}

The baseline content-based model uses  around 5.5 million
features extracted from 
content that the users interacted with. 
These user activities include page views and clicks. Each of these activities is converted to a category represented by a wiki-entity\footnote{\yifan{E.g., if a user read an article about the department store Macy's, a categorical variable {\tt{wiki\_Macy's}} is added to the list of features describing the user}} or an internal Yahoo category.
Then, a logistic regression model is applied on these features to predict gender of an input user.

\section{Models for Gender Prediction}
\label{sec:models}

There are multiple possible approaches to infer a user's gender. A content-based baseline model uses categorical features extracted from content the user interacted with. It has the advantage of identifying users by browser cookies, so works for users who are not registered or not logged in. Its primary disadvantages are that content signals could be sparse, and signals may be lost in the process of distilling into  categorical features. Consequently, the performance is modest (AUC of ROC around 0.80).

Since the goal of this paper is to enhance the gender prediction accuracy for registered users who did not
declare their genders, we propose to use their first names to infer their genders. \revise{We note that while the first name might be a rather random combination of letters, there's still a societal mechanism that makes us interpret certain names as female or male. This means that ad-hoc creation of a name won't work but deciphering this societal gender coding via machine learning can work.

}

Given the first name of a user, there are two approaches to infer the gender. One approach is to use the name embeddings as features. In a prior work, Han et al.~\cite{name_embedding_fakename} \yifan{established ethnic and gender homophily in email correspondence patterns, namely that on aggregate, there is greater than expected concentration of names of the same gender and ethnicity in the contact lists of email users.
Consequently they} constructed name embeddings
by applying word2vec~\cite{Mikolov_2013_word2vec} to the contact lists of millions of users, and used 
the embeddings 
to infer the nationality and cultural origins~\cite{name_embedding_nationality}. We implemented this approach as a second baseline. However, while the embedding based approach works well for most users, word2vec requires a name to appear enough times in contact lists to be able to build meaningful embeddings. For languages such as English, there exists a set of given names that most people adopt. 
\revise{Nevertheless, there are variations of 
common first names that are less often seen. For example,
``Liliana'' is in SSA DATA, but its variation ``Lilianna'' is not in SSA DATA.}
This problem of
rare names is far worse for some character-based languages such as Chinese or Japanese, where a person's first name consists of a few characters in a set of tens of thousands of characters. Thus, first names in these languages are of a very high cardinality and are often unique to a person. Consequently, we found that the embedding-based approach misses many names in its vocabulary set, leading to a degraded performance.

\subsection{Why character-based models?}

\begin{figure*}[htbp]
\includegraphics[
  width=0.48\textwidth,
  keepaspectratio]{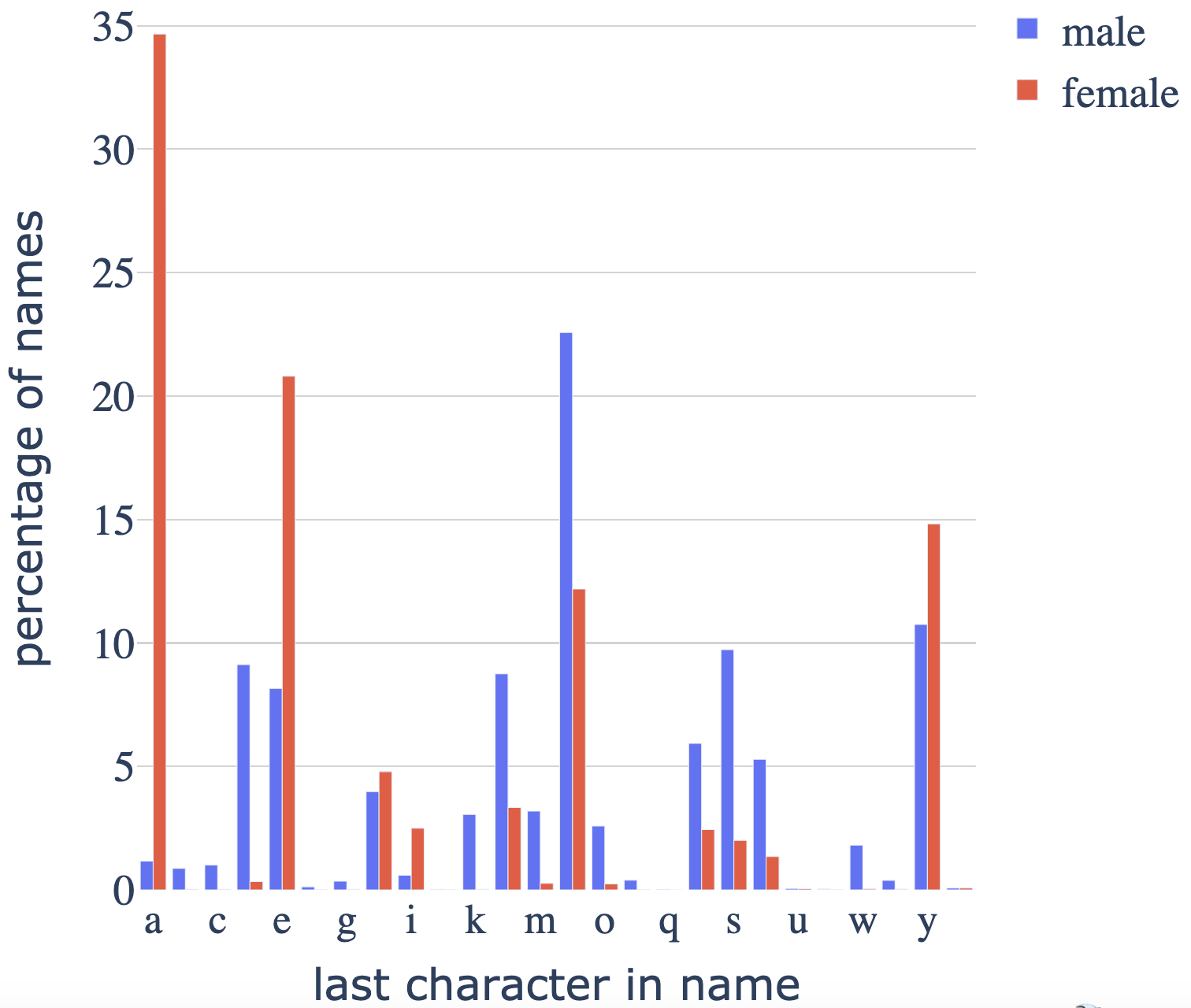}\includegraphics[
  width=0.48\textwidth,
  keepaspectratio]{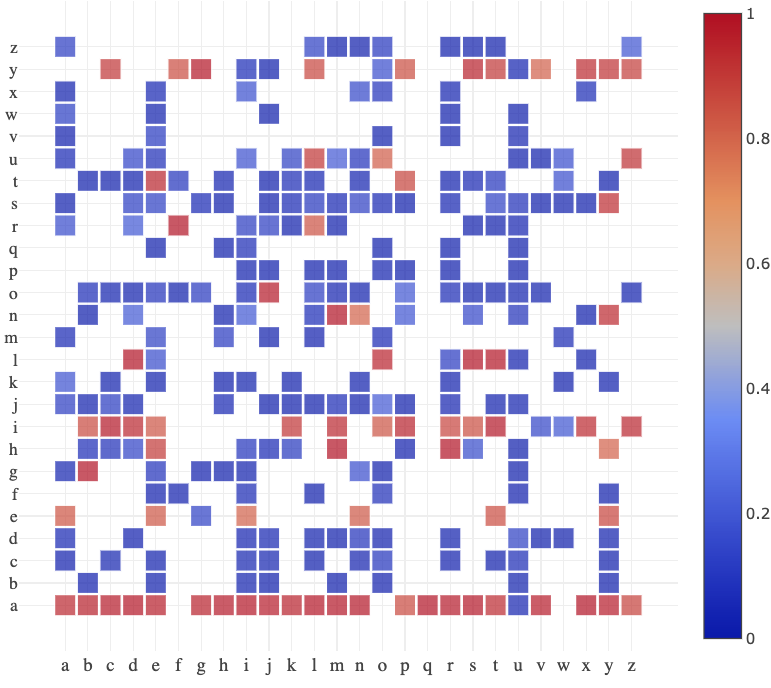}
\caption{Left: distribution plot of the last character in \ssa among \revise{boys} and \revise{girls}. Best viewed in color. Right: heatmap of the female probability, $P(F)$, given the last two characters in \ssa. X-axis is the 2nd last character, y-axis the last character. Only showing character 2-grams with $P(F)>0.7$ or $P(F) < 0.3$, and frequency $\ge 40$.
\label{fig:lastchar}}
\end{figure*}

\begin{CJK*}{UTF8}{gbsn}
Instead of using name embeddings, we propose character-based approaches to infer the gender, because certain characters, or a combination of a few of them, convey the gender of a person. 
For example, in English, first names ending with ``a" (``Linda'', ``Maria'' etc) are more likely female names. Figure~\ref{fig:lastchar}~(left) shows a distribution plot of the last character in names. We can see that
34.7\% of \revise{women} have first names that end with ``a", vs 1\% \revise{of men}. In Chinese, certain characters are only used for female  (e.g., 丽，meaning ``beautiful"), or for male  (e.g., 勇，meaning ``brave").

Beyond single characters, character n-grams can also help to infer genders. For example,
Figure~\ref{fig:lastchar}~(right)
shows a heatmap of $P(F)$ (female 
probability) based on the last two characters of names. For example, names end with ``?a" (last row) are mostly females,
except for ``ua", which 
has a 98.1\% probability of being male (e.g., ``Joshua''). In fact, names where the second last character is ``u" (6th last column) mostly refer to males. Overall, out of 326,765,644 people in \ssa, a large percentage (78.7\%) of them have names that end with a character 2-gram that can be found in Figure~\ref{fig:lastchar}~(right), which implies 
$P(F) \ge 0.7$, or $P(M) \ge 0.7$.

\end{CJK*}

Therefore, \yifan{we believe that character n-grams are correlated with gender.}
This leads us to build character-based models trained from millions of names and their known genders. {\em A key advantage of character-based approaches is that since the features used are characters and character n-grams, the models can work on all names, regardless of whether they are popular or rare.} In particular, it does not suffer from the out-of-vocabulary words issue that the embedding-based approach faces.

In the following, we present a number of character-based models for the gender classification task. We will discuss these models at the algorithmic and architectural level. Implementation details and results will be presented in Section~\ref{sec:results}.

\subsection{Class scaled logistic regression (\nblr)\label{sec:logit}}

An effective yet simple model often used in text classification is NBSVM~\cite{nbsvm}. In this paper we used a similar model, except that we replace the SVM with logistic regression. \revise{The  resulting algorithm
is referred to as class scaled logistic regression, but is denoted as \nblr in contrast to NBSVM}. We describe its  details in Algorithm~\ref{alg:nblr}. The input includes first names and declared gender labels. The model leverages character n-grams, tf-idf and log ratio information for the prediction of gender.

\begin{algorithm} \caption{Class Scaled Logistic Regression (\nblr)} \label{alg:nblr} 
\begin{algorithmic}  
        \STATE $\bullet$ Form character n-grams of the name corpus (we use $n$ between 1 to 7). 
        \STATE $\bullet$ Compute the tf-idf (sparse) matrix $X$. \STATE $\bullet$ Compute the log ratio $r_j$ for each column $j$ of $X$, then scale the column with $r_j$.
    \STATE $\bullet$ Train a logistic regression model using the scaled matrix $X$ and the labels.
\end{algorithmic}
\end{algorithm}

The log ratio $r_j$ of feature $j$ measures the log-odds of the feature. Specifically, let the feature matrix be $X\in R^{m\times n}$, binary labels be $y\in \{0,1\}^m$, then $r_j$ is defined as the logarithm of the ratio between the average value of the elements of column $j$ of $X$ associated with  positive labels in $y$, and the average value associated with negative labels: $    r_j = \log\left(\frac{(1+\sum_{i: y_i = 1} X_{ij})/(1+\sum_{i: y_i = 1} 1)}{(1+\sum_{i: y_i = 0} X_{ij})/(1+\sum_{i: y_i = 0} 1)}\right).
$

\subsection{DNN} 
A fully connected deep neural network (DNN) is a versatile nonlinear model that, given a large set of training data, can learn the weights of hidden layers to minimize the final loss function. We used a DNN with multiple hidden layers. \yifan{As input to the DNN, we first encode each character in a name with a one-hot vector, and then concatenated one-hot vectors associated with each character to a longer binary vector based on the order of their occurrences in the name.}

\subsection{Byte-CNN}

Character-level CNN (char-CNN) has also been applied to natural language processing~\cite{Zhang_cnn_2015}. In char-CNN, each character is represented by a vector of one-hot encoding or fixed-length embedding. These vectors are concatenated into a matrix to
represent a sequence of characters (one or multiple sentences). One dimensional  CNN is applied to this input matrix. In doing so, CNN combines nearby characters and essentially builds up a hierarchy of character grams. Character level encoding is useful for English. But for multilingual text and text with complex symbols like emojis, character level encoding will introduce a lot more parameters. Since our \xobni is multilingual and contain many emojis, we propose {\rm byte-CNN}, a modified version of char-CNN which uses UTF-8 encoding to reduce the number of parameters.

\subsection{LSTM}

\begin{figure*}[t]
\centering
\includegraphics[
  height=0.35\textwidth,
  keepaspectratio]{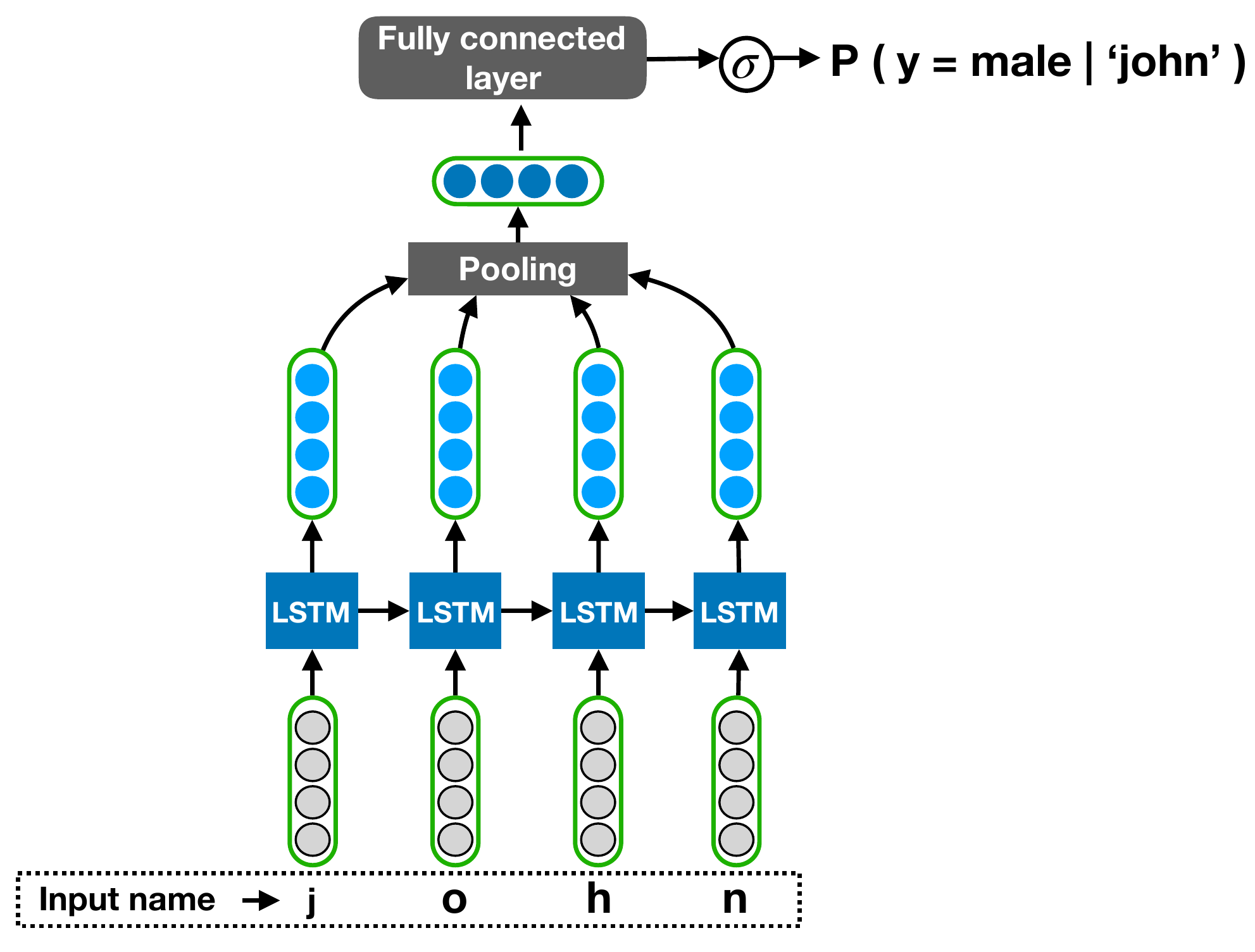}\includegraphics[
  height=0.35\textwidth,
  keepaspectratio]{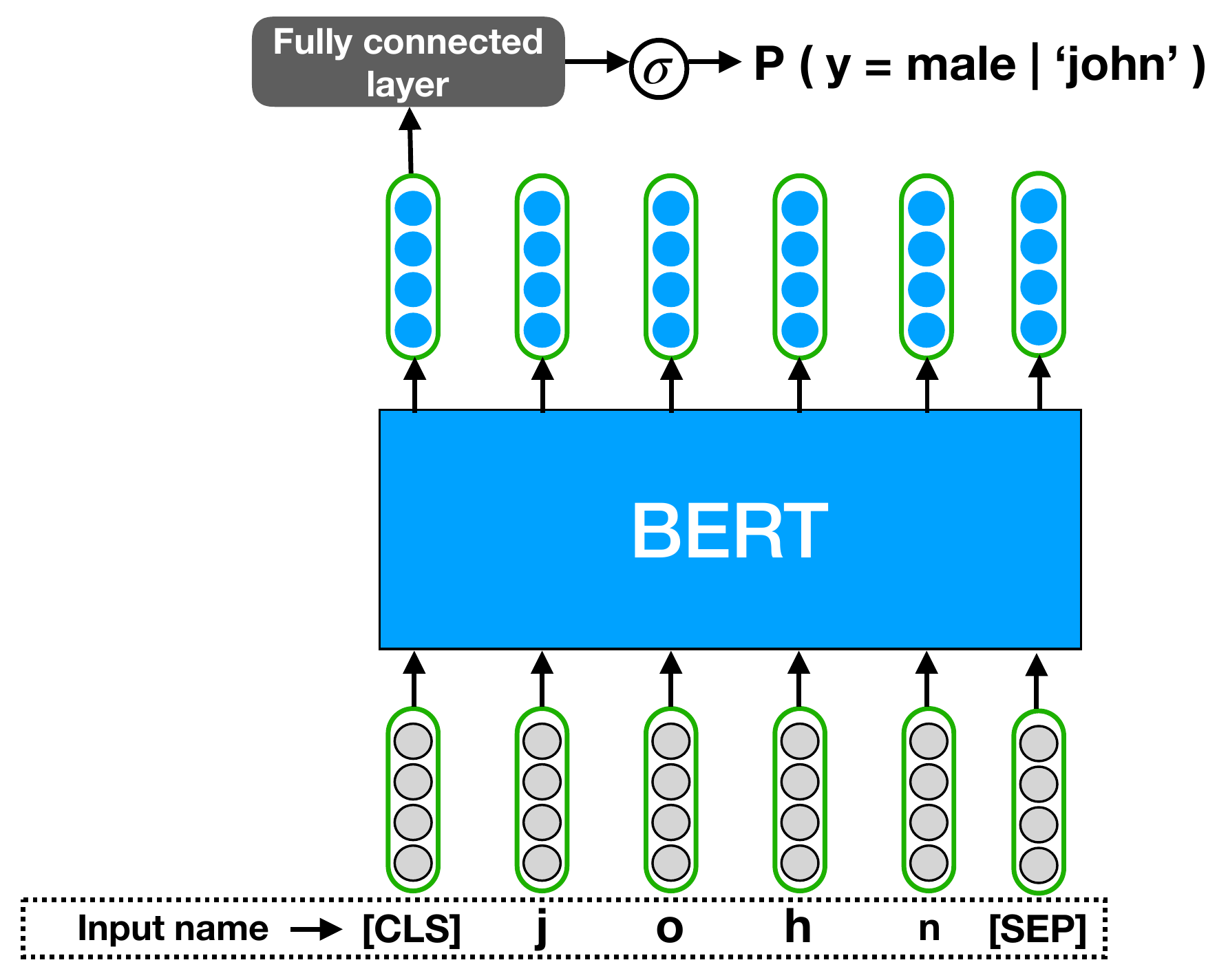}
\caption{Left: character-based LSTM model with pooling. Right: character-based BERT model}
\label{fig:lstm-model}
\end{figure*}

\noindent Long Short-Term Memory (LSTM)~\cite{lstm_hoch_1997} is a recurrent neural network based model that is especially suited for sequence data, in this case characters. In the gender classification task using input names, we consider each character of an input name as a unit and parameterize each character as an embedding vector. Then, we model the transition from characters to characters. Together with a traditional LSTM design, where the output of the last character is used for classification tasks with an intuition that it summarized all information from previous characters, we consider a pooling layer (i.e. Figure \ref{fig:lstm-model}~(left)) to combine results from all outputs of LSTM cells. In this paper, we experiment with \emph{max} and \emph{mean} pooling. 

\subsection{Char-BERT}

\noindent This model originated from a recent state-of-the-art language model \emph{BERT}~\cite{devlin2018bert}. 
Instead of using tokens from WordPiece~\cite{wordpiece_2016}, here the basic units are characters. 
Figure~\ref{fig:lstm-model}~(right) shows the character-based BERT architecture for the gender prediction task with an input name ``John''. Similar to the LSTM model, we encode each character of the input name by an embedding vector. Note that \emph{[CLS]} and \emph{[SEP]} tokens are inserted into the beginning and ending of the input name, analogous to the BERT design in \cite{devlin2018bert}. \emph{[CLS]} and \emph{[SEP]} are also encoded by two separated embedding vectors. Next, the output embedding vector of the \emph{[CLS]} token is used for the gender classification task as it self-attentively aggregates representations of all other tokens in the input name.

\subsection{Name embedding (EMB)\label{sec:emb}}

This is a baseline model using name embeddings~\cite{name_embedding_fakename,name_embedding_nationality}. Embeddings of first and last names are built by the word2vec~\cite{Mikolov_2013_word2vec} model using the contact lists of millions of users as ``sentences". 
They are then used as feature vectors in training and testing.

\section{Experimental Results}
\label{sec:results}

We consider the gender prediction problem as a binary classification problem. Given that some names are used by both \revise{men} and \revise{women}, it may appear at first that this  should be considered as a regression problem\footnote{As an example, in \ssa, around 21\% of people with the name ``Avery" are male. In a regression setting, we can fit a model to predict a value of 0.21 given this name. On the other hand, in a binary classification setting, we seek to predict a value of 0 for this name.}. Even though 10,773 out of 98,440 (11\%) names in \ssa are unisex, our preliminary tests on some of our models show that treating gender prediction as a regression problem does not perform as well as treating it as a binary labelled classification problem. \revise{Hence, for the rest of this paper, we consider gender 
prediction as a binary classification problem.}

\revise{The simplest way to label a name with a binary label is to use majority voting. This is a reasonable strategy especially considering that the \xobni clearly contains noise. For example, names such as ``John" and ``Maria" that are of single-gender in \ssa are not observed to be so in \xobni (Table~\ref{tab:ssa_xobni_top}). Thus,
using the majority voting to decide the gender is helpful in these unambiguous
cases. A more principled way is to treat the problem as one of {\em probabilistic binary classification}. For example, ``Avery'' is documented as male 2098 times and female 8053 times in the \ssa, hence we can have two samples of ``Avery", one with the label of 1 and a sample weight of 2098, the other with 
the label of 0 and a sample weight of 8053. We experimented with this
probabilistic treatment in combination with \nblr. We call the resulting
algorithm \nblrw. We found later in Table~\ref{tab:compare_models} that \nblrw
has similar performance compared with \nblr, indicating that treating 
the gender classification as a pure binary classification problem, 
as opposed to a
probabilistic binary classification problem, is sufficient, with the advantage
of reduced memory and computational requirements.
}

We split the \xobni containing 21M unique names randomly into train, validation and test sets with a ratio of 70/10/20, respectively. We train our models using 
the train set, fine tune model parameters with the validation set, and report results on the \xobni test set, and on the \ssa.
\yifan{Notice that because the size of the \xobni is quite large, the variability when using different random splits is small. For example, when using 5-fold validation on the 21M \xobni, the standard deviation of  all versions of AUC and accuracy (see Section~\ref{sec:metrics}) are below 0.005. Therefore in the following we only report results on the above randomly split data sets.}

\subsection{Performance measurement\label{sec:metrics}}

Even though we build binary classifiers, when measuring performance, each name should not be treated equally. For example, it is more important to predict the genders of popular names correctly, because errors on these names affect more people. Thus, we use two versions of metrics. \revise{The first version treats {\em each name} as a sample. For example, \ssa has 98,400 unique first names, thus 98,400 samples. We denote the 
AUC of ROC and accuracy for this version as \auc and \acc. 

Since popular names represent more people, in real applications it is often more important to predict the gender correctly
for these names. Therefore the second version of the metrics treats {\em each user} as a sample. For example, 
in \ssa, `Avery' was used by 54,330 boys and 125,747 girls. Thus when computing metrics, we split `Avery' into two 
instances (male vs female) that are assigned the same probability $P(M|Avery)$, 
the male one has a sample weight of 54,330, and female one has a sample weight of 125,747. 
We denote the resulting AUC of ROC and accuracy as AUC and ACC.
}

\subsection{ML models and features}

We experimented with six character-based ML algorithms. In the following we
discuss each of these algorithms and its parameter setting.
\revise{All our character based models take into account of the fact that the sequential ordering of the characters in a name is important,
and encode that automatically. The exception is the \nblr algorithm, which has to take the sequential ordering into account both implicitly in the character n-gram formation, and also explicitly (see below).
For Byte-CNN, DNN, LSTM and char-BERT, we limit the max number of characters to 30, padding empty spaces in front if the name is less than 30 characters, or dropping the part of the name beyond 30 characters.}

{\bf \nblr}: this is the class scaled 
logistic regression with $L_1$ regularization. 
For features, we used character $1,2,\ldots,n$-grams. We experimented with different values of $n$, and found that the performance improves as $n$ increases, but the improvement slows when $n \ge 3$. The performance stabilizes when $n\ge 6$. We thus set $n=7$ for the rest of the experiments. Importantly, we found that if we preprocess the name token by prefixing and suffixing each token with ``\_'', the AUC can be improved by around 1-3\%. This is because without this preprocessing, a bigram such as ``ia'' could be the last two characters from ``Julia'', or the middle two characters from ``Brian''. There is no way to differentiate these two cases. With the preprocessing, we know that a 3-gram ``ia\_'' means that the name ends with ``ia''. Incidentally, we realized afterwards that this is similar to how WordPiece~\cite{wordpiece_2016} tokenizes text. The difference
is that WordPiece adds a "\_\_"  to the beginning of a word, but not to the end of a word.

{\bf Byte-CNN}: To handle multilingual names and emojis, we proposed byte-CNN in which we encoded multilingual strings or emojis using UTF-8. UTF-8 is a superset of ASCII, and used one byte to encode the ASCII set for the first 128 characters, which include all \revise{letters}, numbers and regular symbols. For contents that can not be encoded with ASCII codes, UTF-8 uses more than one byte for encoding. For instance, UTF-8 encodes one Chinese character as three bytes. For \ssa, since all names are in English, byte-CNN is equal to char-CNN in this case. For multilingual \xobni, UTF-8 encoding results in 178 unique encoded bytes. We used an embedding layer with 512 dimensions to encode all 178 unique bytes, and then adopted a deep architecture similar to char-CNN, but the embedding is applied on byte level instead of character level. The byte-CNN network includes three dilated convolution layers with 64, 128, and 256 kernels, respectively, \revise{with a kernel size of 3 for all three layers.}

 {\bf DNN}: Similar to byte-CNN, we used UTF-8 to encode multilingual \xobni. A one-hot vector was used to represent a UTF-8 byte, and a fixed number of these one-hot vectors were concatenated into a long vector to represent a name. We experimented with numerous configurations of a fully connected deep neural network. The optimal configuration has 5 layers with corresponding \yifan{1024}, 1024, 512, 256, and 128 hidden nodes, and with a dropout rate of 0.2.

{\bf LSTM}: Given an input name with $l$ characters, an LSTM model will output $l$ embedding vectors. As mentioned before, we processed the $l$ output embedding vectors in 3 strategies: (i) using the last output embedding vector
; (ii) using a \emph{max} pooling layer on top to combine; (iii) using an \emph{average} pooling layer on top to aggregate. We experimented with all three versions, and found that the first version with an embedding size of 192 provided the best results on the validation set. Hence, we only report results from this version.

{\bf char-BERT}: \emph{char-BERT}'s architecture is the same as that of \emph{BERT}, and \emph{char-BERT} can be seen as a pure attention network of only transformation operators without using any language-based pretraining. Note that we can pretrain \emph{char-BERT} with a masked character prediction task, but it is out-of-scope in this paper and we leave it for a future work. For LSTM and char-BERT, we varied the embedding sizes from \{12, 24, 48, 96, 128, 192\}, fixed the learning rate at 0.001, and optimized with Adam optimizer \cite{kingma2014adam}. For \emph{char-BERT} model, we varied the number of attention heads and layers from \{2,3,4,6\}. 
\yifan{We found that \emph{char-BERT} obtained its best results on validation set when using 3 layers with 6 attention heads and an embedding size of 192. We use this setting to get the results reported in the paper.}

{\bf EMB}: \yifan{We used the name embeddings from~\cite{name_embedding_fakename}, which filter} out names that appear less than 10 times in the contact lists. 
Consequently low frequency names in our training and testing data may not have an embedding,
in which case we set them to a vector of zeros. An XGBoost  model was trained using the embedding feature vectors and the labels. 
We also built a logistic regression model and found its performance to be similar. So we did not report its results here.

Note that we do not further compare with other attention-based models such as Attention-based RNN \cite{wang2016attention,chen2017recurrent,zhou2016attention} because char-BERT, which originated from BERT \cite{devlin2018bert}, is a purely attentive transformer network, and has outperformed all previous models.

\subsection{\yifan{Low frequency names for training}}
\label{sec:popular}

We 
consider whether to use all training data to build
models.
Since labels for unpopular names may be unreliable, at first glance 
we should restrict training to popular names with \yifan{high frequency (frequency $\ge \upsilon$)}. To decide the value of  $\upsilon$, we train a \nblr model on \xobni train set, but with name frequency  $\ge \upsilon$. 
Since \ssa can be considered reliably labelled, we use the whole \ssa as a test set for this experiment, filtering
out names that existed in the train data.

Table~\ref{tab:ktrain} shows the \nblr's performance when varying $\upsilon$ in a range of \{1, 2, 5, 10, 20, 40, 80, 100\}.
Clearly, it improves as we lower the frequency
threshold, and using all available training data
gives the best test performance on \ssa. This indicates
that even though individual labels for low frequency names
may be unreliable,
collectively, these names still provide valuable signals that improve classifier performance.
Therefore, we proposed to use all
available training data for the rest of this paper.

\begin{table}[htbp]
    \centering
    \begin{tabular}{c|c|c|c|c}
$\upsilon$&\auc&\aucww&\acc&\accww\\
\hline
1&0.958&0.991  & 0.897&0.963 \\
2&0.951&0.990   & 0.885&0.957   \\
5&0.940&0.986   & 0.874&0.944   \\
10&0.930&0.98   & 0.862&0.934  \\
20&0.921&0.972 & 0.853&0.920   \\
40&0.914&0.962 & 0.846&0.907  \\
80&0.906&0.945 & 0.837&0.882   \\
100&0.902&0.940 & 0.833&0.880  \\
\end{tabular}
    \caption{Performance of \nblr models trained on \xobni with name frequency $\ge \upsilon$ and tested on the whole \ssa.}
    \label{tab:ktrain}
\end{table}

\subsection{Comparing models on \xobni}

\revise{For the rest of the paper, we denote names with name frequency  $k < \kk$ as unpopular names, for reasons explained in 
Section~\ref{sec:dict}.} We first compare the performance of
different models on \xobni test data \revise{with these unpopular names.}
The results are given in Table~\ref{tab:compare_models} (top). EMB -- the embedding-based baseline model  performed poorly, because the embedding for many of these low frequency names do not exist. 
DNN, while being a more complicated model, actually performed worse than the linear model \nblr. The best performance comes from the \nblr,  LSTM and char-BERT models, with \aucww of 0.872/0.872/0.874, respectively.
These three models are very similar in performance across all metrics. 

\begin{table}[htb]
    \centering
    \begin{tabular}{c|c|c|c|c}
model&\auc&\aucww&\acc&\accww\\
\hline
 & \multicolumn{4}{c}{on test data with name frequency $k < 40$}\\
\hline
\nblr      &{\bf 0.879}&{\bf 0.872}&{\bf 0.799}&{\bf 0.797}\\
\nblrw      & 0.859 & 0.856 & 0.783 &  0.785 \\
DNN       & 0.850  & 0.851 & 0.775 &  0.780 \\
Byte-CNN & 0.838 & 0.836 & 0.765 &  0.767 \\
LSTM      & {\bf 0.879}  & {\bf 0.872} & {\bf 0.799}  & {\bf 0.798} \\
char-BERT & {\bf 0.881}  & {\bf 0.874} & {\bf 0.801}  & {\bf 0.799} \\
EMB       & 0.717  & 0.724 & 0.676  & 0.685 \\
\hline
 & \multicolumn{4}{c}{All test data}\\
\hline
\nblr      &{\bf 0.879}&{\bf 0.940}&{\bf 0.800}&{\bf 0.876}\\
\nblrw      &0.860&0.935&0.785&0.873\\
DNN       & 0.851  & 0.903 & 0.776 & 0.837\\
Byte-CNN  & 0.839  & 0.889 & 0.766 & 0.814 \\
LSTM      & {\bf 0.880}  & {\bf 0.938} & {\bf 0.801} & {\bf 0.871}\\
char-BERT & {\bf 0.883}  & {\bf 0.935} & {\bf 0.803} & {\bf 0.871}\\
EMB       &{0.722}&{0.927}&{0.678}&{0.854}\\
CONTENT& -     &0.800&  -   &  0.729\\
\end{tabular}
    \caption{Top: performance for different ML models, trained with all \xobni training data, tested on testing data with name frequency $<40$. Bottom: tested on all testing data. }
    \label{tab:compare_models}
\end{table}

The bottom of Table~\ref{tab:compare_models}  shows the performance of the CONTENT baseline on all test data (e.g. regardless of the name frequency). 
CONTENT uses features extracted from the content the users interacted with, as discussed in Section \ref{sec:content}. The \aucww is 0.800, with \accww of 0.729. For comparison, 
\nblr has \aucww of 0.940, with corresponding accuracy of 0.876, both much higher than the CONTENT baseline. LSTM and char-BERT also performed well.
The other baseline model EMB performs better than the CONTENT model, with \aucww on all test data of 0.920 and corresponding accuracy of 0.855, though inferior to \nblr, LSTM or char-BERT. \revise{
\nblrw, which treats the gender 
prediction problem as a probabilistic binary classification problem,
has similar or slightly worse performance compared with \nblr.}

We also checked the names on which our models do not work well. We define the loss of a prediction on a name as $\left(P(M|\text{first}) - 0.5\right)(|F| - |M|)$
where $|M|$ and $|F|$ are the number of \revise{men} and \revise{women} with that name. Names with the highest losses 
are those that are relatively popular, and where the prediction is 
wrong. The top 10 names with the highest loss for \nblr are shown in Table~\ref{nblr_bad}. Most of them are names commonly used by both \revise{men} or \revise{women}, such as ``Robin''. One exception is ``Negar'', which is a female name of Persian origin. This indicates that inference for unisex names is a challenging task. In Section~\ref{sec:fullname} and Section~\ref{sec:name_and_content} we offers two
possible enhancement.

\begin{table*}[htb]
    \centering
        \setlength{\tabcolsep}{4.5pt}
    \begin{tabular}{c|c|c|c|c|c|c|c|c|c|c}
  & Robin & Sina & Jan & Justine & Micah & Negar & Zia & Val & Asal & Lane\\
\hline
M/(M+F) & 0.38 & 0.8 & 0.46 & 0.37 & 0.70 & 0.05 & 0.79 & 0.37 & 0.14 & 0.77\\
P(M) & 0.74 & 0.16 & 0.94 & 0.72 & 0.25 & 0.72 & 0.17 & 0.84 & 0.81 & 0.28
\end{tabular}
    \caption{Names with the highest prediction loss. M/(M+F) is the observed male probability.}
    \label{nblr_bad}
\end{table*}

\subsection{Generalizability of the models}

In this section, we are interested in understanding whether
models trained on the large scale \xobni name data
are generalizable. In addition, we would like to find out if models trained
using the smaller \ssa data are extendable as well.

For these experiments, we split \ssa into an 80/20 ratio, corresponding to
train/test sets. To test the 
generalizability properly, we took out all names
in \ssa test data that existed in \xobni train set, leading to 2,077 remaining names in \ssa test set.

Table~\ref{tab:compare_models_ssa}
gives the results of our five models. We trained using the train sets for \xobni and \ssa, and tested
on the test sets for \xobni and \ssa, respectively.

\begin{table}[htb]
    \centering
\begin{tabular}{c|c|c|c|c|c|c}
\multicolumn{3}{c|}{} & \multicolumn{2}{c|}{any $k$} & \multicolumn{2}{c}{$k < 40$}\\
\hline
model & train & test & \aucww & \accww & \aucww & \accww\\
\hline
\nblr & Yahoo & SSA & 0.972 & 0.916 & 0.963 & 0.905\\
DNN & Yahoo & SSA & 0.970 & 0.926 & 0.962 & 0.907 \\
Byte-CNN & Yahoo & SSA & 0.955 & 0.901 & 0.949 & 0.888 \\
LSTM & Yahoo & SSA & {\bf 0.980} & {\bf 0.940} & {\bf 0.971} & {\bf 0.925} \\
char-BERT & Yahoo & SSA & {\bf 0.980} & 0.930 & {\bf 0.971} & 0.921  \\
\hline
\nblr & SSA & SSA & {\bf 0.986} & {\bf 0.947} & { 0.977} & { 0.939} \\
DNN & SSA & SSA & 0.978 & 0.938 & 0.951 & 0.897\\
Byte-CNN & SSA & SSA & 0.971 & 0.926 & 0.940 & 0.877\\
LSTM & SSA & SSA & {\bf 0.985} & {\bf 0.953} & {\bf 0.992} & {\bf 0.960} \\
char-BERT & SSA & SSA & 0.955 & 0.891 & 0.982 & 0.936 \\
\hline
\nblr & SSA & Yahoo & 0.859 & 0.791 & \textbf{0.756} & \textbf{0.710} \\
DNN & SSA & Yahoo & \textbf{0.874} & \textbf{0.823} & 0.700 & 0.674 \\
Byte-CNN & SSA & Yahoo &  0.852 & 0.790  & 0.687 & 0.663\\
LSTM & SSA & Yahoo & 0.871 & 0.812 & 0.710 & 0.682\\
char-BERT & SSA & Yahoo & 0.849 &  0.797 & 0.683 & 0.668 \\
\hline
\end{tabular}
    \caption{Testing the generalizability of the models trained on \xobni or \ssa.} 
    \label{tab:compare_models_ssa}
\end{table}

{\bf Training on \xobni and testing on \ssa}: from Table~\ref{tab:compare_models_ssa}~(top),
we see that LSTM, char-BERT and \nblr generally work the best, and their performance \yifan{is} consistent with their results reported in Table~\ref{tab:compare_models}. \nblr (\aucww$=0.972$) is slightly worse than LSTM and char-BERT (\aucww$=0.980$). 
However the differences are very small, indicating that \nblr works effectively for predicting user's genders. Importantly, all our models perform very well on SSA test data, indicating that {\em our models trained on \xobni generalize well to \ssa}.

{\bf Training on \ssa and testing on \ssa}: from Table~\ref{tab:compare_models_ssa}~(middle), we see that when trained and tested using the \ssa, \nblr and LSTM  perform the best, but others also work reasonably well, indicating that \ssa, as an English only dataset, is homogeneous and relatively easy to predict.

{\bf Training on \ssa and testing on \xobni}: In Table~\ref{tab:compare_models_ssa}~(bottom), we build models using the smaller \ssa train set and evaluate on large-scale \xobni test set. All models perform worse compared to their corresponding performance reported in Table \ref{tab:compare_models} (i.e. trained on \xobni train data and evaluated on \xobni test data).  
This indicates that \emph{building models using the smaller \ssa train data does not generalize well to the larger \xobni at all}. For example,
\nblr only achieves \aucww of 0.859 on the \xobni test set, vs \nblr trained using \xobni, which achieved \aucww of 0.940 in Table~\ref{tab:compare_models}. This is not
a limitation of our models, but due to the fact that \ssa is an English based dataset, so
models trained using it could not be expected to understand the non-English/multilingual names in \xobni. \revise{We note that another reason that \xobni seems to be harder
to predict than \ssa is due to the presence of noise. The absence of noise in the latter also makes it easier to classify.}

Overall, we conclude that models trained using the larger \xobni extend very well to names that are not observed in the train data.

\revise{
\section{Using Both First and Last Names For Gender Classification\label{sec:fullname}}

While the gender is typically conveyed in the first name, there are cases when the same first name can be used for one gender in one culture, but another in a different culture. For example, the Italian male name ``Andrea'' (derived from the Greek ``Andreas'') is considered a female name in many languages, such as English, German, Hungarian, Czech, and Spanish~\cite{wiki:unisex}. Since the last name often 
conveys a person's ethnicity, we speculate that it may be
beneficial to include it in the gender classification for such unisex names. In this section
we attempts to answer this conjecture.

\subsection{A Dual-LSTM model that utilizes  both the first and the last names}

\begin{figure}[htbp]
\centering
\includegraphics[width=0.4\textwidth, keepaspectratio]{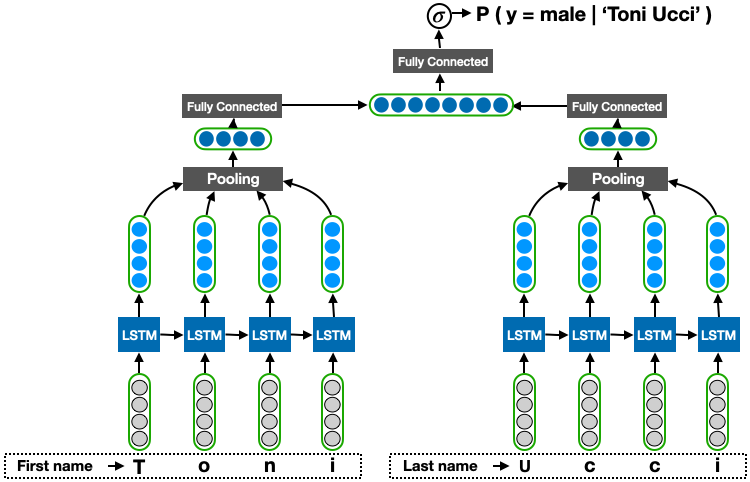}
\caption{The dual-LSTM (DLSTM) model that utilizes both  first and last names}
\label{fig:dlstm-model}
\end{figure}

While the linear classifier \nblr was found to be simple yet effective for predicting gender using first names, it is not suited for utilizing full names. One way to include the last name in a linear model would be to construct features that conjunct character-based n-gram features for both first and last names. However, doing so would increase the feature cardinality quadratically. Therefore it is necessary to explore different approaches.

We propose to encode each of the first and last names into a dense vector via character LSTM, then  predict the gender using a concatenation of the two vectors. We call this the dual-LSTM model, or DLSTM for short. Figure~\ref{fig:dlstm-model}
illustrates the model architecture.

For comparison, we propose another approach using
name embeddings~\cite{name_embedding_fakename,name_embedding_nationality} via
 the EMB algorithm. As explained in Section~\ref{sec:emb}, 
both the first and last name are embedded into a 100-D space by applying word2vec to the contact lists of millions of users. We then encode a full name into a
200-D vector by concatenating their respective embeddings, and apply XGBoost to predict the gender.

\subsection{Results of models using both first and last names}
\label{sec:fullname_results}

To conduct this experiment, we used \xobniB, which contains the first and last names of 13M users.
 We split the data into 70\% train, 10\% validation and 20\% test sets,  
making sure that the same full name only appears in one of these 3 sets. We also experimented with
 a 5-fold cross validation, and found that
the variance for \aucww and \accww were less than 0.001. Therefore, we only report the mean results.

For DLSTM, each character is embedded as a 256 dimensional vector,
before passing through LSTM units. We experimented with adding an attention layer in place of the pooling layer, but did not find it helpful. Instead, as in LSTM, we find that using the embeddings of the last characters work well. We fed the last embeddings to a dense layer of 300 neurons, then concatenated the output into a 600-D vector.
We found that it is necessary to have fully connected layers after the concatenation, in order to 
allow the two embeddings to interact properly.  We used fully connected layers with 512, 256, 64 neurons, before feeding to a sigmoid function that gives the final predicted probability. We limited the max characters 
to 30 per arm of the DLSTM, and pad empty space in front for short names. To make a consistent comparison, the LSTM
model in this section follows a similar setup (we find that it performed similarly to the LSTM in the previous sections).

Table~\ref{tab:lastname} shows the results of
the LSTM and DLSTM classifiers. 
For comparison, we also give the corresponding results when using the name embedding algorithm EMB. 
It is seen that DLSTM that uses both the first and last names improves the \aucww/\accww by 0.8\%/1.1\% respectively,
when compared with LSTM that uses first name only, indicating that information from the last name is helpful. 
LSTM model also works better than the EMB model. 

As an additional test, we also tried a simpler way of utilizing the full names: we
use the first and last names as a single string separated by a space (shown as ``LSTM/full string'' in Table~\ref{tab:lastname}), and applying the LSTM algorithm (Figure~\ref{fig:lstm-model}~left). This simpler way of utilizing first and last names does not perform as well as
DLSTM, even though it is a little better than LSTM using first name alone 
(shown as ``LSTM/first'' in Table~\ref{tab:lastname}). 

For a fair comparison, we also applied \nblr  to first names.  We give its performance in Table~\ref{tab:lastname}.  It does perform similarly to LSTM with first names, but as discussed earlier it is not suitable for taking advantage of both first and last names. It is also notable that both LSTM and \nblr give better performance on this full name dataset 
than on \xobni (Table~\ref{tab:compare_models}). This is  because \xobniB is sampled based on full names, as such the first names in this data set tend to be more popular. Consider the population of users in \xobni, 
we found that first names in \xobni have an average of only 20 users, vs 704 for \xobniB. We knew from Table~\ref{tab:compare_models} that it is easier to classify the gender of popular names. 
We did try applying \nblr to full names by concatenating first and last names with a special character separator, the result is even worse than using first names alone.

\begin{table*}[htb]
    \centering
    \setlength{\tabcolsep}{2.2pt}
    \begin{tabular}{c|c|c|c|c|c|c|c|}
 & & \multicolumn{2}{c|}{\xobniB} & \multicolumn{2}{c|}{ \andrea}&\multicolumn{2}{c|}{ \toni}\\
\hline
model & features  & \aucww & \accww & \aucww & \accww & \aucww & \accww\\
\hline
LSTM   & first & 0.949 & 0.889 & 0.509 & 0.648 & 0.506 & 0.556 \\
LSTM  & full string & 0.953 & 0.895 & 0.841 & {\bf 0.827} & 0.826 & 0.792\\
DLSTM  & first$+$last & {\bf 0.957} & {\bf 0.901} & 0.823 & 0.796 & {\bf 0.886} & {\bf 0.850}\\
EMB  & first & 0.941 & 0.871 & 0.5 & 0.506 & 0.5 & 0.648\\
EMB  & first$+$last & 0.946 & 0.879 & {\bf 0.857} & 0.783 & 0.854 & 0.811\\
\nblr & first & 0.951 & 0.891 & 0.5 & 0.556 & 0.5 & 0.352\\
\hline
\end{tabular}
    \caption{Performance of DLSTM and EMB (with XGBoost) using both first and last names, vs LSTM, \nblr and EMB with only first names. Testing is on the test set of the \xobniB, or on the \andrea and the \toni. 
    } 
    \label{tab:lastname}
\end{table*}

}

\revise{

We wanted to see if
the LSTM model trained using both the first and the last names was especially effective for unisex first names. 
We knew that unisex first names are not as common as gendered first names. For example, about 11\% of the Social Security baby names~\cite{ssa_babies} are documented as both male and female.
Given the relative small percentage of unisex names,
the effect of a model on these names may be diluted when looking at the overall performance in Table~\ref{tab:lastname}.

We divide users 
in \xobniB into 40 buckets based observed $P(M|\text{first})$
(first name appears less than 10 times are filtered out). We denote these buckets $b[i],\ (i = 1,2,\ldots, 40)$. Specifically, 
$b[i] = [0.025*(i-1), 0.025*i).$
The first few buckets are for people with first names that are  used almost exclusively by women. E.g.,
$b[1] = [0,0.025)$  contains people with a first name that are used by men only $<2.5\%$ of the time. The last few buckets are people with first names that are used almost exclusively by men. E.g., $b[40] = [0.975, 1.0]$
contains people with a first name that are used by men  $\ge 97.5\%$ of the time. The middle buckets contain people with unisex names. For example, $b[20]$ and $b[21]$ contains people with first names used by men between $47.5\%$ to $52.5\%$ of the time. 

Figure~\ref{fig:unisex}~(right) shows the distribution of
people in the 40 buckets. In all plots in Figure~\ref{fig:unisex}, the x-axis, from left to right, corresponds to buckets 1 to 40. Clearly
few people belongs to the unisex buckets.
Figure~\ref{fig:unisex}~(left/middle) shows the \aucww/\accww comparison of LSTM that uses first name only, vs DLSTM 
that uses both first and last names, on each of the 40 buckets. 
As can be seen from Figure~\ref{fig:unisex}~(middle), for the first and last few buckets, which 
correspond to people with first names that are 
exclusively used by women or by men, the \accww difference
between these two algorithms are relatively small.
The gap widens to about 10\% as we moved to the middle area that contains unisex names. When looking at \aucww, LSTM has \aucww values around 50\%, indicating that
whatever the percentage of women/men a bucket has, using first name alone 
is not enough to tell them apart. 
DLSTM, which uses both first and last names,
consistently outperform with 10-15\% higher \aucww.
This confirms  that using full names is beneficial, especially when classifying the gender of unisex names.

\begin{figure*}[htb]
    \centering
\includegraphics[width=0.34\textwidth,keepaspectratio]{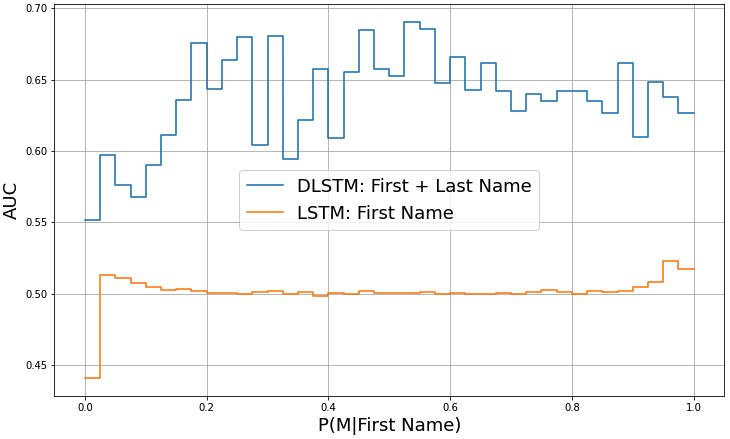}\includegraphics[width=0.34\textwidth,keepaspectratio]{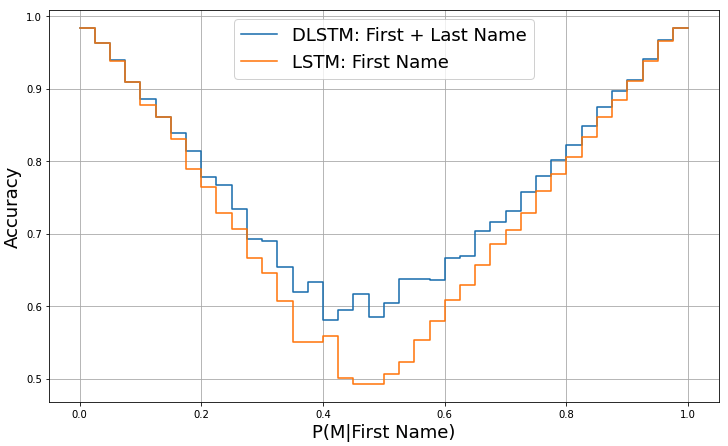}\includegraphics[width=0.34\textwidth,keepaspectratio]{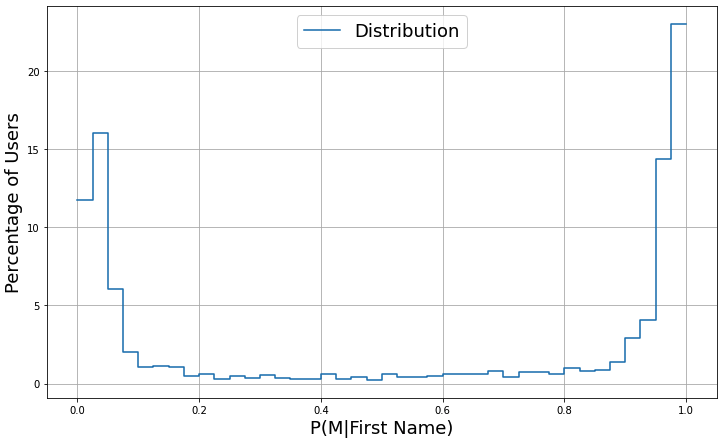}
\caption{The \aucww (left) and \accww (middle) of DLSTM vs LSTM, as a function of how unisex the first names are. People are divided into 40 buckets on the
x-axis based on $P(\text{M}|\text{first})$. Names in the middle of the x-axis, away from 
0 and 1, are unisex names. The distribution of people
are show on the right.\label{fig:unisex}}
\end{figure*}

To given some concrete examples, we collected two unisex name datasets. We found
133 famous people (74 \revise{women}, 59 \revise{men}) with the first name ``Andrea''~\cite{wiki:andrea}, and call this the \andrea. In addition we found 54 famous people (19 \revise{women}, 35 \revise{men}) with the first name ``Toni''~\cite{wiki:toni}, and call this the \toni.
About 16\% of ``Andrea'' and 33\% of ``Toni'' in \xobniB are men.
To avoid overfitting, we removed any person in \xobniB whose full name matches those 187 names in \andrea and \toni.
Table~\ref{tab:lastname} shows that on these two
datasets, the DLSTM model is much more effective
than the LSTM model with only first names. On the \andrea, AUC improved from 0.509 to 0.823, while on the \toni, 
AUC improved from 0.506 to 0.886. 

Here we give some examples where the full name based model works well. For \revise{men}, 
``Andrea Bianchi" (former Italian film director and writer), 
``Andrea Boattini" (Italian astronomer), 
``Andrea Bargnani" (Italian former professional basketball player); 
for \revise{women}, 
``Andrea Sestini Hlaváčková" (Czech professional tennis player), 
``Andrea Růžičková" (Slovak actress and model), 
``Andrea Murez" (Israeli-American Olympic swimmer).

There are   cases where
the model failed, including, for \revise{men}  misclassified as \revise{women}, 
``Andrea Tacquet" (Belgium Mathematician) and
``Andrea Costa" (Italian Master Mason and socialist activist); 
and for \revise{women} misclassified as \revise{men}, 
``Andrea Gyarmati" (Hungarian swimmer) and
``Andrea Fioroni" (Argentine field hockey player). Given that ``Fioroni" is a family name originated from Italy, this last  failure can actually be considered an anomaly in the data. 

}

\section{Related Work}
\label{sec:related}

Given the importance of gender information for ad targeting, unsurprisingly there are many existing works on gender classification.

\subsection{\lastcut{Content based gender classification}}

Grbovic et al.~\cite{Grbovic:2015} studied gender based ad targeting in Tumblr. Their model created gender labels using \ssa as the ground truth. It then uses blog content (titles, texts) and user actions (follows, likes and reblogs) as features, much like the content model described in Section~\ref{sec:content}.

Many works infer the user gender by extracting features based on user activity and user-generated content on social media \cite{kokkos2014robust,ciot2013gender,rao2010detecting,pennacchiotti2011machine,otterbacher2010inferring,burger2011discriminating,liu2012using,al2012homophily,culotta2015predicting,culotta2016predicting,merler2015you,sakaki2014twitter,beretta2015interactive,ludu2014inferring}. Otterbacher~\cite{otterbacher2010inferring} inferred gender of 21k movie reviewers by their writing style, content and metadata. Ciot \textit{et al}~\cite{ciot2013gender} inferred gender of 8.6k Twitter users by the top-k discriminative language features between male and female, such as the top-k hashtags, words, mentions, \emph{etc.} Liu \textit{et al}~\cite{liu2012using} used Twitter content to infer gender of 33k users. Merler \textit{et al}~\cite{merler2015you} used pictures coming from the user's feeds to infer gender of 10k Twitter users. Sakaki \textit{et al}~\cite{sakaki2014twitter} combined both user-generated text and images in their feeds to infer gender of 3.9k Twitter users.

Unlike these works, we only infer users' gender using their  names. This allows us to immediately infer the gender of newly registered users without requiring any user activity or content, and avoids the issue of sparse signals in content-based models.

\subsection{\lastcut{Name based demographic classification}}

Several works predicted gender based on users' names~\cite{knowles2016demographer,mueller2016gender}. 
Liu and Ruths~\cite{LiuR13} 
proposed two methods of incorporating
names and other Twitter features for predicting gender of Twitter users. They found that using the gender association score of a name as a switch to
determine whether to bypass the content-based classifier works better than using both kinds of features. Gender association score was computed using  name distributions from the
1990 US census. However, no name classifier was involved.

Ambekar~\textit{et al}~\cite{ambekar2009name} used full names to infer ethnicity, using hidden Markov models. 
Han \textit{et al.}~\cite{name_embedding_fakename} found evidence of ethnic/gender homophily in email correspondence patterns. Through large-scale analysis of
contact lists, they found that there was a greater than
expected concentration of names of the same gender and
ethnicity within the same contact lists.
Consequently, they generated name embeddings using word2vec, and used the embeddings for gender, ethnicity and nationality classification with good results~\cite{name_embedding_fakename,name_embedding_nationality}.

Brown~\cite{brown2017} proposed a naive Bayes classifier to classify the gender of a name.
The features used are simpler than ours and came from an {\tt nltk} book: first/last letter, count of letters, has letters, suffixes (last 2, 3, 4 letters of name). They also compared with an RNN character-based model and found that the Naive Bayes model performs much better. Compared with their method, on the same dataset~\cite{brown2017}, we found our \nblr classifier to be more accurate (0.878 vs 0.85 ACC.)

Overall, our work is different from previous works in five areas: (1) We studied the gender prediction task on two much larger scale datasets:  the \xobni with 21M unique first names and the \ssa with 98.4k unique names; (2) we built many character-based models, from the \nblr linear model to the advanced char-BERT model, and conduct a comprehensive analysis on these models; (3) we verified the effectiveness of our proposal on multilingual names and its generalizability for unseen datasets; (4) we compared with content and embedding based baselines and showed that our approach is more accurate; (5) \yifan{we demonstrated that the performance of character-based machine learning models could be further enhanced when utilizing both the first and last names, as well as the content the user interacted with (see Appendix). }

\hide{
count	20.000000
mean	0.878422
std	0.001764
min	0.875061
255075max	0.882827

# convert Ellis Brown dataset
df = pd.read_csv("name_gender_dataset.csv", header=None)
def m(s, g):
    if s == g:
        return 1
    else:
        return 0
df["M"] = [m(s, "male") for s in df[1]]
df["F"] = [m(s, "female") for s in df[1]]
df["ratio"] = df["M"]/(df["M"] + df["F"])
df["ratio_binary"] = round(df["ratio"])
df.columns = ["name","MF","F","M","ratio","ratio_binary"]
df = df.drop("MF", axis = 1)
df.to_csv("/tmp/name_gender_dataset_allis_brown.tsv", sep = "\t")
}

\section{Conclusion}
\label{sec:conc}

In this paper, we considered the problem of inferring gender from users' first names.
We proposed character-based machine learning models, and demonstrated that our models are able to infer the genders of users with much higher accuracy than models based on content information, or based on name embeddings. In addition, we find surprisingly that a linear model, with careful feature engineering, performed just as well as complex models such as LSTM or char-BERT. We verify the generalizability of our proposal for unseen datasets.

First names may have different gender connotations in different cultures. 
\revise{We  demonstrated in Section~\ref{sec:fullname} using a dual-LSTM model that utilizing both first and last names can be effective in such a situation.}

Finally, while a content based model offers inferior 
performance to the name based models, we believe that it could be complementary to the latter,
especially for people with unisex names. We conducted a study in Section~\ref{sec:name_and_content} 
showing that an ensemble model that combines both first name and content information 
 gives a further performance boost. 
 
 \revise{One limitation of using names as the main signal for classifying genders is that users may not disclose their true names. Another limitation is that gender may be difficult to infer if the first name is unisex. We showed
 separately that it is helpful to use
  the first and last names, 
 or  the first name and content information. For future work, we intend to investigate whether combining these all together (first and last names, and content information) can
 give us even better performance.}

\section*{Acknowledgment} The authors like to thank the reviewers for the very thoughtful reviews
of  earlier versions of this paper.

\bibliographystyle{spmpsci}

\newpage

\setcounter{section}{0}
\renewcommand\thesection{\Alph{section}}

\revise{

\section{Appendix\label{sec:appendix}}

\subsection{Fusion model that combines content information and names for gender prediction\label{sec:name_and_content}}

We have seen in Section~\ref{sec:results} (Table~\ref{tab:compare_models}) that the content based model 
is inferior to the name based models. However, it is reasonable to
expect that because content consumption by users captures behavior signals, it may offer useful information that
can complement the name based models. This is especially true if the user has a unisex first name, for which it is not possible 
to predict with certainty the gender of the user by first name alone.

We conducted a study to investigate a simple ensemble model. We took a sample of 1M users from \xobniB. 
For each user, we predicted the male probability using the first name and the \nblr model. For the same user, we predicted the probability based on the content the user consumed, using the CONTENT model.

\begin{figure}[htbp]
\begin{center}
\includegraphics[
  width=0.4\textwidth,
  keepaspectratio]{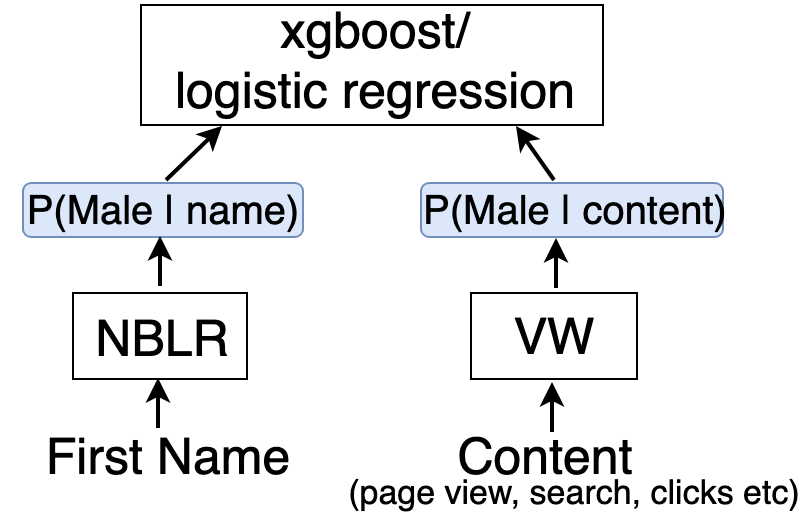}
\caption{Fusion model that combines name and content information to predict gender.}
\label{fig:fusion-model}
\end{center}
\end{figure}

We experimented with the following ways of fusing the results of the content and name based models:

{\bf Logistic regression}: we use  $P(\text{M}|\text{first})$ and $P(\text{M}|\text{content})$ as features and declared gender as labels to fit a logistic regression model.

{\bf Logistic regression with logits}: we first convert  $P(\text{M}|\text{first})$ and $P(\text{M}|\text{content})$ to logits, then fit a logistic regression model. Specifically, we convert the probability to
logits via the function $p\rightarrow \ln(p/(1-p))$.

{\bf Xgboost}: we use  $P(\text{M}|\text{first})$ and $P(\text{M}|\text{content})$ as features and declared gender as labels to fit an xgboost model.

\begin{table}[htb]
    \centering
    \setlength{\tabcolsep}{2.2pt}
    \begin{tabular}{|c|c|c|c|}
\hline
model  & feature & AUC & ACC\\
\hline
logistic regression & first name+content & \underline{0.9764}  & \underline{0.9333} \\
logistic regression with logit &first name+content &  0.9762  & 0.9319 \\
xgboost & first name+content & {\bf 0.9784}  & {\bf 0.9346} \\
 name (\nblr)& first name  & 0.9709  & 0.9230 \\
content &content &  0.7937  & 0.7270 \\
\hline
\end{tabular}
    \caption{Performance comparison of 3 fusion models, vs the baseline name and content models
    } 
    \label{tab_fusion}
\end{table}

We did a 5-fold cross-validation to derive the average results
in Table~\ref{tab_fusion}. We do not report the standard deviation since it is quite small (under 0.0006). From the table, we see that the fusion model using xgboost can improve AUC by 0.008, accuracy by 0.012, confirming our belief that combining the
content and names improves gender prediction. Logistic regression  performed slightly worse than xgboost. Logistic regression with logits does not perform as well as
logistic regression, although still better than the name based model alone.
}

\subsection{Dictionary look-up vs model prediction \label{sec:dict}}

For users with popular names, it is unnecessary to use a name-based machine learning model to predict their genders. Instead, a simple
dictionary lookup is sufficient -- it avoids any error that might be introduced in the modeling  process. On the other hand, for unpopular names, a model-based method is more suitable: when a name 
appears only a few times, statistically we have a low confidence in deciding the gender based on the few occurrences. Furthermore, for names that have never appeared before in the data, a machine learning model is the only choice.

Therefore, when deploying in real world applications, we suggest taking a hierarchical approach for the gender classification: if a name appears $\ge k$ times
in the population of users with declared gender, we decide the gender based on a dictionary lookup. For other names, we
apply the machine learning based models. We
will reason on a suitable choice of $k$ next.

In general, we want to pick a name frequency $k$, above which we can trust the labels in the lookup dictionary with a high confidence. Suppose that a name is 30\%/70\% male/female in a very large population (e.g., the world). If there are $k$ of our users with that name, drawn from that population randomly, then based on simple calculation using binomial distribution,
we need at least $k
\ge 16$ to be sure that the name will be labelled as female with $95\%$ or higher probability~\footnote{\yifan{Given a predominantly
female name with a male probability of $p < 0.5$, and 
given $k$ randomly selected people with this name, the probability that more than half of these people are male is:
$f(p, k) = \sum_{i=\lceil k/2\rceil)}^{k} C_n^i p^i(1-p)^{k-i}$. For $p = 0.3$, we found that $f(p,k) < 0.05$ when $k \ge 16.$ For $p=0.4$, we found that $f(p,k) < 0.05$ when $k \ge 66.$}}. For a name that is 40\%/60\% male/female, we need $k\ge 66$. Based on the above considerations, we decide to choose $k = \kk.$ With that name frequency, we know from Table~\ref{tab:freq_coverage} that a dictionary lookup can cover the gender of 86\% of users. The gender for remaining users can be inferred by 
the machine learning models. {\em We emphasize that
dictionary lookup is proposed only for real-world applications. For the purpose of understanding our models for names of different frequencies, all results reported in this paper are based on the machine learning models.} We have seen in Table~\ref{tab:compare_models} that our proposed models performed
well for both popular and unpopular names, when compared with the baselines.

\subsection{Understanding the models}

In the previous subsection, we have seen that \nblr, LSTM and char-BERT all perform very similarly. \yifan{Normally, a linear model such as \nblr is easier to understand than nonlinear models. By looking at the weights for
the features, one can understand why a sample is predicted as positive or negative.}

However, in the case of predicting the gender using the name string, the features are character n-grams extracted from the same name, hence they are not independent. 
This feature correlation can give rise to negative weights
for some features that are associated with male names, which is
compensated by larger positive weights for other correlated features (\revise{as a reminder, we use label 1 for male and 0 for female}).
Similarly positive weights may be observed for features that are associated with female names. This makes model interpretation from feature weights difficult.
For example, when looking at the female name ``\_anna\_'' (we add prefix and suffix ``\_'' to a name, see Section~\ref{sec:logit}), we found that the 5-gram ``anna\_'' has a positive weight of 0.36, which may be a surprise as one would expect that a name ends with ``anna'' should be a female name. However, a substring of ``anna\_'' is ``nna\_'', and it has a negative weight of $-1.19$. When combining the weights of all 17 character grams, together with the intercept, we end up with a weight of $-9.01$ and a male probability of close to 0. Thus, looking at feature weights can be confusing and even misleading.

Instead, it is more instructive to apply \yifan{one of our models} to real datasets, and see how it behaves.
We first apply \nblr to classify Chinese names. Note that the Chinese language
has up to 100K characters. Of these, about 7,000 are commonly used. A person's given name could consist of any one or two characters. Therefore, the number of unique given names
could be in the millions ($7,000\times7,000=49\text{M}$). It is difficult to collect enough instances of each of these possible variations and their associated gender. This is where a character-based model is especially suited.
As a sanity check, we looked at
a list of 11 names from~\cite{eleven_chinese_names}
that are considered especially beautiful, because of
their poetic connotation or phonetic harmony. 
Table~\ref{tab:chinese_names} shows the male probabilities
for each of these names, together with a ``note'' column for their likely genders and explanation, 
taken from~\cite{eleven_chinese_names}.
We observe
that our classifier does assign higher scores for names
that are definitely male than those that are female.

\begin{CJK*}{UTF8}{gbsn}

\begin{table}[htp]
    \centering
    \setlength{\tabcolsep}{3.5pt}
    \begin{tabular}{c|c|c|l}
name & $P(M)$ & label & note\\
\hline
芷若(F)&0.00&F  & Name for girls of two herbal plants\\
语嫣&0.00&F & Popular female character by Jin Yong\\
映月&0.04&F & Name for girls. ``reflection of the moon''\\
念真(F)&0.12&M/F & A belief in truthfulness\\
徽因&0.26&F & Famous female Chinese poet/architect\\
望舒&0.29&M/F & God who drives the moon carriage\\
如是&0.30&F & Famous Chinese courtesan and poet\\
莫愁&0.64&M/F & Free of sadness\\
风眠&0.77&M/F & Falling asleep in the woods/breeze\\
青山(M)&0.85&M & Blue-coloured mountains\\
飞鸿&0.87&M & A swan goose soaring high in the sky\\
\end{tabular}
    \caption{``Eleven Beautiful Chinese Names'' from~\cite{eleven_chinese_names}. $P(M)$ is the probability of male, predicted by \nblr. The ``label"  column gives the likely gender, and
    the ``note'' column gives a description, both based on~\cite{eleven_chinese_names}. Three of the eleven has frequency $\ge 40$ and their genders
    (in brackets) could have been looked up.
        \label{tab:chinese_names}
    }
\end{table}
\end{CJK*}

It is also helpful to understand the \nblr classifier by looking at English words that scored high as male or female. We first take the top 3,000 most frequently used English words\yifan{~\cite{english3000}}, filter out 443 words that are found in \ssa (such as ``Rose''), then list the top 11 highest scored male/female
words in Table~\ref{tab:top3000} (a). Interestingly, a lot of them make sense, e.g., ``miss'' or ``women'' are
female words due to their meanings. However, we notice that the majority
of these 3,000 words are found in \xobni as names. Next, we take 236,736 words in {\tt nltk.corpus.words} and filter out any names in either \ssa or \xobni, and end up with 169,902 words. 
The highest scoring masculine/feminine words are given in Table~\ref{tab:top3000} (b). Many of the words 
in column ``feminine'' of Table~\ref{tab:top3000} (b) do look
very feminine. For example, many words end in ``ia'' (e.g., ``anadenia''). 
\yifan{Words on the right-hand-side of the table are classified as masculine, likely due to the ending of the words, such as ``*boy''. Incidentally ``knappan" is a word in Malayalam (a Dravidian language primarily spoken in the south Indian state of Kerala) 
that means ``a good for nothing guy".}

\begin{table}[htb]\centering
    \begin{tabular}{l|l|l|l}
\multicolumn{2}{c|}{(a) Top 3,000 words}&\multicolumn{2}{c}{(b) {\tt nltk.corpus.words}}\\
\hline
feminine & masculine & feminine & masculine\\
\hline
miss&gentleman& Myrianida&unclement\\
woman&father& nonylene&farcial\\
Mrs&engineer& decylene&maccoboy\\
makeup&commander& shamianah&comradery\\
criteria&military& whidah&knappan\\
guideline&hard& Manettia&subramous\\
nurse&big& anadenia&herdboy\\
pregnancy&auto& mollienisia&beshod\\
\yifan{mom}&\yifan{dad}& \yifan{Anacanthini}&\yifan{Geophilus}\\
\yifan{daughter}&\yifan{kill}& \yifan{phenylene}&\yifan{clocksmith}\\
\yifan{wife}&\yifan{electronic}&\yifan{vinylene}&\yifan{Bavius}\\
\hline
\multicolumn{4}{c}{(c) S\&P 500 companies}\\
\hline
\multicolumn{2}{c|}{feminine}&\multicolumn{2}{c}{masculine}\\
\hline
\multicolumn{2}{l|}{Tiffany \& Co.}&\multicolumn{2}{l}{Emerson Electric}\\
\multicolumn{2}{l|}{Ulta Beauty	}&\multicolumn{2}{l}{Cisco Systems}\\
\multicolumn{2}{l|}{Kimberly-Clark	}&\multicolumn{2}{l}{Philip Morris International}\\
\multicolumn{2}{l|}{DaVita Inc.	}&\multicolumn{2}{l}{Raymond James Finance Inc.}\\
\multicolumn{2}{l|}{Apple Inc.}&\multicolumn{2}{l}{Henry Schein}\\
\multicolumn{2}{l|}{Nike}&\multicolumn{2}{l}{Mohawk Industries}\\
\multicolumn{2}{l|}{Danaher Corp.}&\multicolumn{2}{l}{Duke Energy}\\
\multicolumn{2}{l|}{CIGNA Corp.}&\multicolumn{2}{l}{General Motor}\\
\multicolumn{2}{l|}{\yifan{Lilly (Eli) \& Co.}}&\multicolumn{2}{l}{\yifan{Stanley Black \& Decker}}\\
\multicolumn{2}{l|}{\yifan{Macy's Inc.}}&\multicolumn{2}{l}{\yifan{Edison Int'l}}\\
\multicolumn{2}{l|}{\yifan{Dentsply Sirona}}&\multicolumn{2}{l}{\yifan{General Electric}}\\
\hline
\end{tabular}
    \caption{Words with a high female/male score among: (a) 3,000 most frequent English words that are not in \ssa; (b) 236,736 words in {\tt nltk.corpus.words} that are neither in \ssa nor in \xobni. (c) S\&P 500 companies.}
    \label{tab:top3000}
\end{table}

Table~\ref{tab:top3000} (c) gives the most 
feminine and masculine company names in S\&P\ 500 companies~\cite{sp500}. The results are also \yifan{interesting. Corporations with feminine names include a number of ladies fashion companies (Tiffany, Ulta, Macy's), while
many with masculine names are in engineering or energy sectors (e.g., Duke Energy, General Motor, General Electric).}

Overall, the performance of \nblr on these three datasets illustrates that the model can extend well
to rare or even unseen strings, and to non-English
languages.

\end{document}